\documentclass[sigconf]{acmart}

\usepackage{booktabs, multirow, amsmath}
\usepackage[ruled]{algorithm2e}
\usepackage{url}

\usepackage{flushend}

\DeclareMathOperator*{\argmax}{arg\,max}
\DeclareMathOperator*{\softmax}{softmax}
\DeclareMathOperator*{\sigmoid}{sigmoid}

\copyrightyear{2019}
\acmYear{2019} 
\setcopyright{iw3c2w3}
\acmConference[WWW '19]{Proceedings of the 2019 World Wide Web Conference}{May 13--17, 2019}{San Francisco, CA, USA}
\acmBooktitle{ Proceedings of the 2019 World Wide Web Conference (WWW'19), May 13--17, 2019, San Francisco, CA, USA}
\acmPrice{}
\acmDOI{10.1145/3308558.3313491}
\acmISBN{978-1-4503-6674-8/19/05}

\newcommand{\eat}[1]{}

\newcommand{\ie}{\textit{i.e.,}\xspace}
\newcommand{\eg}{\textit{e.g.,}\xspace}
\newcommand{\etal}{\textit{et al.}\xspace}

\newcommand{\paratitle}[1]{\noindent\textbf{#1}}

\begin{document}
\title[DLocRL: Deep Pipeline for Twitter Location Recognition and Linking]{DLocRL: A Deep Learning Pipeline for Fine-Grained Location Recognition and Linking in Tweets}

\newcommand{\ourApproach}{\textsc{DLocRL}\xspace}

\author{Canwen Xu}
\orcid{0000-0002-1552-999X}
\affiliation{%
  \institution{Wuhan University}
  \department{School of Computer Science}
  \state{Hubei}
  \country{China}
}
\email{xucanwen@whu.edu.cn}

\author{Jing Li}
\affiliation{%
  \institution{Inception Institute of Artificial Intelligence}
  \country{United Arab Emirates}
}
\email{jing.li@inceptioniai.org}

\author{Xiangyang Luo}
\authornote{Corresponding author.}
\affiliation{%
  \institution{State Key Lab of Mathematical Engineering and Advanced Computing}
  \state{Henan}
  \country{China}
}
\email{xiangyangluo@126.com}

\author{Jiaxin Pei}
\affiliation{%
  \institution{Wuhan University}
  \department{School of Computer Science}
  \state{Hubei}
  \country{China}
}
\email{pedropei@whu.edu.cn}

\author{Chenliang Li}
\affiliation{%
  \institution{Wuhan University}
  \department{School of Cyber Science and Engineering}
  \state{Hubei}
  \country{China}
}
\email{cllee@whu.edu.cn}

\author{Donghong Ji}
\affiliation{%
  \institution{Wuhan University}
  \department{School of Cyber Science and Engineering}
  \state{Hubei}
  \country{China}
}
\email{dhji@whu.edu.cn}

\begin{abstract}
	
In recent years, with the prevalence of social media and smart devices, people causally reveal their locations such as shops, hotels, and restaurants in their tweets.  
Recognizing and linking such fine-grained location mentions to well-defined location profiles are beneficial for retrieval and recommendation systems. In this paper, we propose \textbf{\ourApproach}, a new deep learning pipeline for fine-grained location recognition and linking in tweets, and verify its effectiveness on a real-world Twitter dataset.
\end{abstract}




\maketitle
\section{Introduction}\label{sec:intro}

Twitter is a place where users can share their daily life activities by posting tweets (\ie short messages, up to 140 characters each). 
In many tweets, locations are implicitly or causally revealed by users at fine-grained granularity \cite{sigir14:li,www16:ji,jasist17:li}, for example, a restaurant, a shopping mall, a park or a landmark building. 
Here, a fine-grained location is equivalent to a point-of-interest (POI), which is a focused geographical entity \cite{sigir14:li,rae2012mining}.
In this paper, we target on \textit{recognizing} mentions of POIs and \textit{linking} these mentions to well-defined location profiles. 

The two tasks are important for several reasons.
First, recognizing mentions of POIs is beneficial for information retrieval \cite{balog2018entity}.   
An example is that query understanding can be enhanced by exploiting POI information in location-based information retrieval systems \cite{espinoza2001geonotes,schiller2004location}. 
In addition, recognized POIs can be integrated into knowledge bases and support many business intelligence applications such as POI recommendation and location-aware advertising \cite{www13:lingad,sigir14:li,oh2003effects}. 
Second, the linked location profile can serve as side information for the tweet, and vice versa. 
For example, Twitter sentiment analysis can be conducted in a more precise manner by incorporating the location profile content \cite{han2018linking}.

However, recognizing POI mentions and linking these mentions to well-defined location profiles are both challenging.
Due to the informal writing of tweets, POIs are usually mentioned by incomplete name, nickname, acronym or misspellings. 
 For example, the mention \textit{vivo} may refer to \textit{VivoCity} which is a comprehensive shopping mall in Singapore, or a smartphone brand. 
Even if we have successfully recognized a POI mention, it remains challenging to link the mention to a specified location profile (\ie a specific shopping mall with address and geo-coordinates in this case).

On the other hand, existing solutions \cite{han2018linking,www16:ji,sigir14:li} on these two tasks require a large set of features manually designed for each task and domain, which demands task and domain expertise. 
For example, Li \etal \cite{sigir14:li} carefully designed lexical, grammatical, geographical and BILOU schema features (totally 11 features) to extract fine-grained locations from tweets. 
Ji \etal \cite{www16:ji} proposed a joint model to recognize and link fine-grained locations from tweets with 24 hand-crafted features.  
Recently, Han \etal \cite{han2018linking} proposed a probabilistic model with seven types of hand-crafted features to link fine-grained location in user comments. 
 It is now generally admitted that distributed representations could better capture lexical semantics \cite{goldberg2017neural}. 
We would envision for a system that is based on distributed representations, and that can learn informative features for location recognition and linking by itself without human effort.

In this paper, we propose \ourApproach, a new \textbf{D}eep pipeline for fine-grained \textbf{Loc}ation \textbf{R}ecognition and \textbf{L}inking in tweets. 
\ourApproach is designed to adopt effective representation learning, semantic composition, and attention and gate mechanisms to exploit the multiple semantic context features for location recognition and linking. 
\ourApproach consists of two core modules: \textit{recognition} module and \textit{linking} module.

The \textit{recognition} module aims to extract a text segment referring to a fine-grained location (\ie POI) from a given tweet.
In DLocRL, we formulate the recognition task as a sequence labeling problem (\ie assigning tags to tokens in the given sequence).
We adopt bi-directional long short-term memory with conditional random fields output layer (BiLSTM-CRF) to train a POI recognizer.

The \textit{linking} module aims to link each recognized POI to a corresponding location profile. 
Before linking POIs to location profiles,  an extensive collection of location profiles (CLP) is constructed.  
Given an input pair $\langle$POI, Profile$\rangle$, the \textit{linking} module is trained to judge whether the location profile corresponds to the POI. 
More specifically, \ourApproach utilizes two pairs of parallel LSTMs to encode the left context and right context for a POI, respectively. 
The location profiles are represented by their constituents with TF-IDF weights and one-hot schema. 
The Manhattan distance is used to measure the geographical distances between users and profiles.
Finally, the representation of  $\langle$POI, Profile$\rangle$ comes from three sources: tweet-level contextual information, location profile representation and geographical distance. 
The representation is then fed into a fully-connected layer for final linking prediction. 
Moreover, to take advantage of the geographical coherence information
among mentioned POIs in the same tweet, we develop Geographical Pair Linking (Geo-PL), a post-processing strategy to further enhance the linking accuracy.

In summary, we make the following contributions: (1) We propose \textbf{\ourApproach}, a new deep learning pipeline for fine-grained Location Recognition and Linking in tweets.  To the best of our knowledge, our work is the first attempt to address location recognition and linking in the paradigm of deep learning. (2) We develop the Geographical Pair Linking (Geo-PL) approach, a post-processing strategy to further improve linking performance. (3) We conduct extensive experiments on a real-world Twitter dataset. The experimental results show the effectiveness of \ourApproach on fine-grained location recognition and linking.  We also conduct ablation studies to validate the effectiveness of each design choice.

\section{Related work}
\label{sec:related}
\subsection{Location Mention Recognition and Disambiguation}
\paratitle{Recognition.}
Facing noisy and short tweets, traditional NER methods suffer from their unreliable linguistic features \cite{tkde18:zheng}. 
Prior solutions exploit comprehensive linguistic features like Part-of-Speech tags, capitalization \cite{conll09:ratinov}, Brown clustering \cite{coling92:brown} to improve recognition performance. 
For example, Li \etal \cite{sigir14:li} carefully designed lexical, grammatical, geographical and BILOU schema features (totally 11 features) to extract fine-grained locations from tweets. 
Ji \etal \cite{www16:ji} proposed a joint model to recognize and link fine-grained locations from tweets with 24 hand-crafted features. 
Zhang \etal\cite{josis14:zhang} designed four types of features and trained a classifier to select the best match among gazetteer candidates. 
Generally, location gazetteers are widely used in location mention recognition \cite{pacling15:malmasi,josis14:zhang,sigir14:li}.
For noisy text, Malmasi \etal\cite{pacling15:malmasi} proposed an approach based on Noun Phrase extraction and n-gram based matching instead of the traditional methods using NER or CRF. 
Different from these existing works, our approach \ourApproach does not require hand-crafted features. 

\paratitle{Disambiguation.}
For entity disambiguation, many studies attempt to exploit the coherence among mentioned entities. 
Pair-wise \cite{kdd09:kulkarni} and global collective methods \cite{emnlp11:hoffart,sigir11:han,jmlr05:tsochantaridis} have been applied to explore the coherence.
Zhang \etal\cite{josis14:zhang} and Ji \etal\cite{www16:ji} observed that the geographical coherence is effective in location disambiguation task.
Also, instead of tweet-level coherence, Li \etal\cite{icde14:li} utilized user-level coherence to facilitate entity disambiguation. 
Besides, Shen \etal \cite{kdd13:shen} introduced user interest modeling to collective disambiguation.

To form an entirely end-to-end model for joint recognition and linking, Guo \etal\cite{naacl13:guo} and Ji \etal\cite{www16:ji} adopted structural SVM and perceptron, respectively. 
Furthermore, the beam search algorithm \cite{acl08:zhang} is also used in \cite{www16:ji} to search the best combination of the mention recognition and linking.
In this work, we devise \ourApproach, a new pipeline which can be upgraded and optimized much easier than previous joint models and still achieves better performance.

\subsection{Neural Networks for NER and EL}

\paratitle{Named Entity Recognition (NER).}
The use of neural models for NER was pioneered by Collobert \etal \cite{collobert2011natural}, where an architecture based on temporal convolutional neural networks (CNNs) over  word sequence was proposed.
Chiu \etal ~\cite{chiu2015named} utilized CNN to detect character-level features and LSTM to capture the word-level context in a sentence.
Yang \etal ~\cite{yang2016multi} proposed a gated recurrent unit (GRU) network to learn useful morphological representation from the character sequence of a word.
Recently, Peters \etal~\cite{peters2018deep} proposed deep contextualized word representation, which can model syntax, semantics, and polysemy.
Akbik \etal ~\cite{akbik2018contextual} proposed contextual string embeddings, to leverage the internal states of a trained character language model to produce a novel type of word embedding. 

\paratitle{Entity Linking (EL).}
Neural networks were firstly used for entity linking by Sun \etal\cite{ijcai15:sun}.
They proposed a model which takes consideration of the semantic representations of mention, context, and entity.
Recently, Phan \textit{et al.} \cite{cikm17:phan} proposed a neural network for entity linking with LSTM and attention mechanism.
They also proposed \textit{Pair Linking} to enhance collective linking by measuring the cosine similarity of the text embeddings between two mentioned entities.

\section{Model Description}

\begin{figure}
	\includegraphics[width=8.5cm]{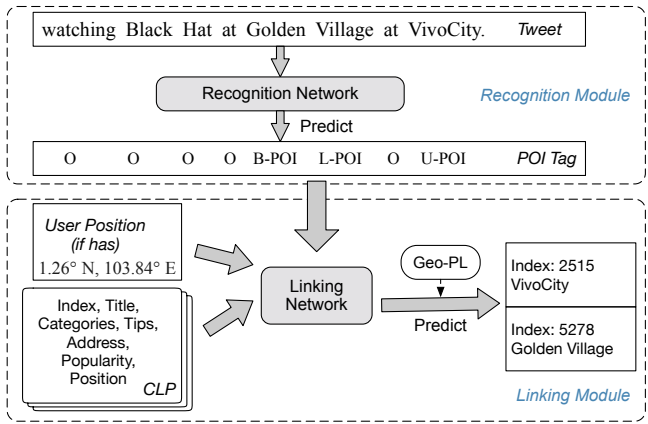}
	\caption{A running example to illustrate the workflow of \ourApproach.}
	\label{fig:arch}
\end{figure}

Figure~\ref{fig:arch} shows the workflow of \ourApproach, which consists of recognition module and location linking module.

\subsection{Location Recognition}
\label{subsec:recognition}

\begin{figure}
	\includegraphics[width=8.5cm]{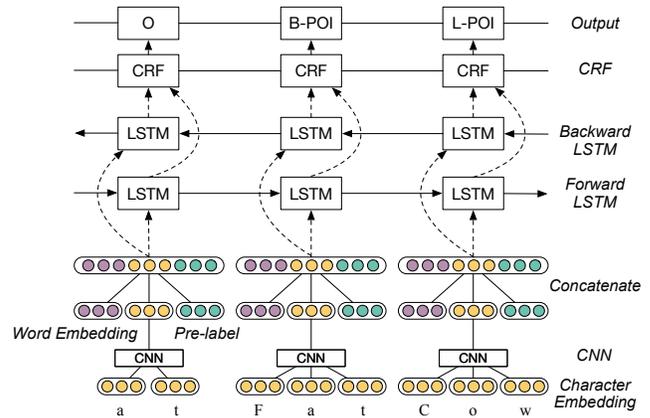}
	\caption{The architecture of location recognition module.}
	\label{fig:recognition}
\end{figure}

Figure \ref{fig:recognition} shows the architecture of the location recognition module. 
The representation of a word consists of its pre-trained word embedding, BILOU pre-label and character-level representation. 
Finally, BiLSTM-CRF is utilized to infer tag sequence based on the representations. 

\paratitle{Pre-trained Word Embedding.}
To deal with the problem of informal spellings and casual expressions, we should capture the information from tweets as accurate as possible.
We use the GloVe\cite{emnlp14:pennington} embeddings pre-trained on a large-scale Twitter corpus of two billion tweets.
The vocabulary of pre-trained embeddings covers most common misspellings and aliases of most common words.

\paratitle{Pre-label.}
POI inventory, containing partial and familiar names from Foursquare, is proposed in \cite{sigir14:li} to pre-label candidate location mentions in tweets.
It has been proved that pre-label is an essential resource to improve recognition performance.
We use a CRF toolkit\footnote{CRF++: https://github.com/taku910/crfpp} to automatically assign the pre-labels with BILOU scheme\cite{conll09:ratinov}.

\paratitle{Character-level Representation.}
Previous studies~\cite{chiu2015named,huang2015bidirectional} have shown that character-level information (\eg prefix and suffix of a word) is an effective resource for NER task.
CNN and BiLSTM are commonly used in previous works to extract character-level representation.
In our model, we use CNN because of its lower computational cost \cite{chiu2015named,ijcai2018-579}. 

\paratitle{BiLSTM-CRF.}
LSTM is a variant of recurrent neural network (RNN) and is designed to deal with vanishing gradients problem. 
BiLSTM uses two LSTMs to represent each token of the sequence based on both the past and the future context of the token. 
As shown in Figure \ref{fig:recognition}, one LSTM processes the sequence from left to right, the other one from right to left.
For a word, we concatenate its pre-trained word embedding, pre-label and character-level representation as its final representation, which is then fed into a BiLSTM layer. 
Then, the output sequence is fed into a CRF layer to infer the tag sequence. 
BiLSTM-CRF is a state-of-the-art approach to named entity recognition \cite{huang2015bidirectional,acl15:dyer}.

%Shown in Section \ref{subsec:overallcmp}, our model gets a good \textit{recall} which benefits the linking subtask since a location profile mapping dictionary (discussed in Section \ref{subsec:linking}) is capable of filtering out the false positive (FP) predictions in the later process.

\subsection{Location Linking}
\label{subsec:linking}

\begin{figure*}
	\includegraphics[width=18cm]{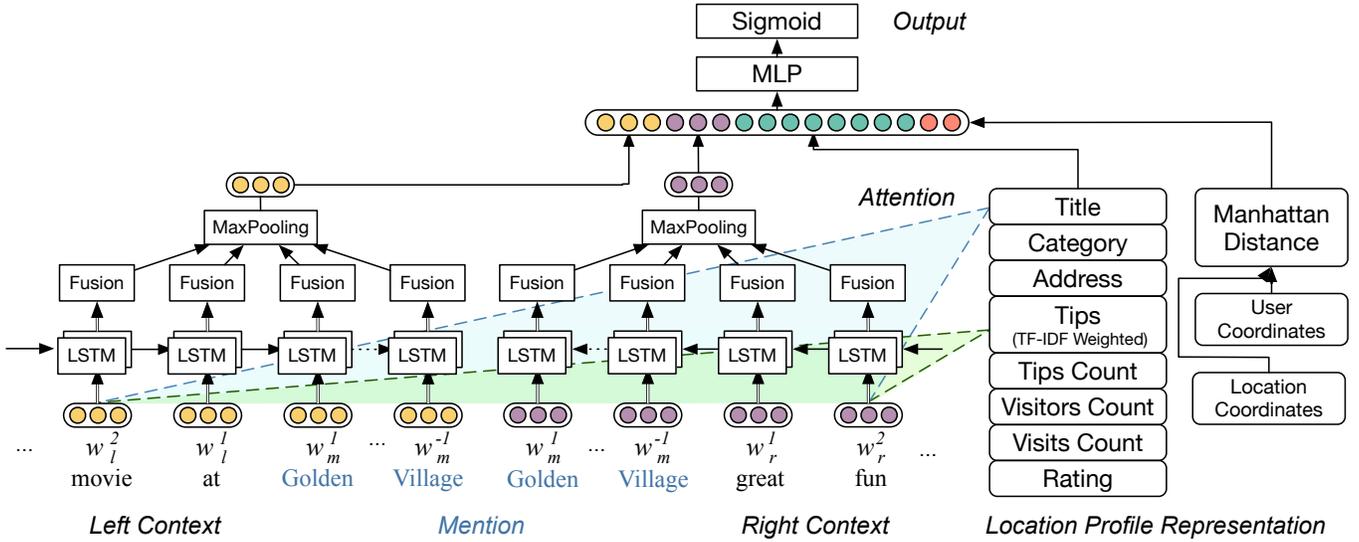}
	\caption{The architecture of location linking network. The input is a pair $\langle$tweet, profile$\rangle$, and the output is the matching score.}
	\label{fig:linking}
\end{figure*}

 Figure \ref{fig:linking} shows the architecture of the location linking module. 
We use two pairs of parallel LSTMs (four in total) to encode the left-side and right-side contexts of a mention, respectively.
Note that the input of the two parallel LSTMs for each side is re-weighted by two different \textit{attentions} from the location profile.
The output sequences of LSTMs are ``fused'' by a \textit{fusion gate} and then fed into a max-pooling layer over all time
steps as the final representation for each side context. 
Next, we concatenate the left-side context representation, right-side context representation, the representation of location profile together 
with the \textit{Manhattan distance} between user coordinates (\ie user position attached to the tweet) and coordinates from location profile. 
Finally, a multilayer perceptron (MLP) with a sigmoid activated output layer is used to output a scalar ranged between zero and one as the matching score.

\paratitle{Location Profile Mapping Dictionary.}
Following \cite{www16:ji}, we use a location profile mapping dictionary to recall all possible candidates for a location mention.
The \textit{key} of the dictionary is a possible POI mention, and the \textit{value} is the list of candidate profile indexes for the POI mention.
If the mention is not in the mapping dictionary, we predict it as \textit{un-linkable}. 
The dictionary is constructed with Foursquare check-in tweets.

\paratitle{Mention's Context.}
Following \cite{cikm17:phan}, we use LSTM networks to capture two-side contextual information of the POI mention. 
The difference is that we use all words in a tweet instead of specific window size. 
The left-side context starts from the leftmost word in a tweet and ends at the rightmost word inside the mention. 
Conversely, the right-side context starts from the tweet's rightmost word and ends at the leftmost word of the mention. 
Since the tweet does not have a long context, we use all words to understand the tweet as a whole. 
Note that the input of LSTM layers is re-weighted with a multi-attention mechanism, which is to be detailed shortly.

\paratitle{Behavioral and Semantic Information.}
Users often reveal their locations and describe what they are doing in tweets.
For example, a user posts  ``Great lobster @ Red Robin!'', which contains information about behavior (having lobster).
Here, we name such information as \textbf{\textit{behavioral information}}.
Obviously, since few restaurants serve lobster, it could be beneficial for POI disambiguation. 
According to our observation, tweets share the same behavioral information with the tips (\ie user comments) in the Foursquare location profile.
On the other hand, some words in a tweet may be directly semantically relevant to the category or address of a location.
For instance, in the tweet ``Best wine at Elle's in the center'', \textit{wine} is closer to \textit{bar} in word embedding space. 
Such information is beneficial for location disambiguation between \textit{Elle's Bar} and \textit{Elle's Salon}. 
Similarly, in the same example, \textit{center} is a synonym for \textit{Central Region}.
This can be helpful for picking a location profile titled ``Elle's Bar - Central Region'' out of other branches of Elle's Bar.
We name such information as \textbf{\textit{semantic information}}.

\paratitle{Multi-attention Mechanism.} To comprehensively exploit \textit{semantic information} and \textit{behavioral information}, we 
develop a multi-attention mechanism to assign weights for the input sequence. 
As shown in Figure \ref{fig:linking}, we use the \textit{title} representation and \textit{tips} representation as two attention vectors. 
Specifically, given an input word embedding $w_i$ in a tweet and an attention vector $p$
(\ie representation of title or tips from location profile), the 
re-weighted input word embedding $w'_i$ is defined by:
\begin{equation}z_i = \tanh{(V_pp + V_ww_i)}\end{equation}
\begin{equation}s_i = \softmax(v_kz_i+v_b)\end{equation}
\begin{equation}w'_i = w_is_i\end{equation}
where $V_*$, $v_k$, and $v_b$ are attention parameters which can be learned during training. 
The TF-IDF weighted tips representation (to be detailed shortly) contains the most characteristic behavioral information about a location. 
Thus, the attention from tips can highlight the parts which are most relevant with the tips.
Similarly, the title contains rich semantic information (including \textit{location name}, \textit{branch name} and
\textit{business status}) which can highlight the parts relevant to the location.

\paratitle{Fusion Gate.} To collect important information from the output of two parallel LSTMs, we borrow a simple but practical \textit{fusion gate} from \cite{aaai18:shen}.
Given the output of two parallel LSTMs with different attentions (\ie $x^1_i$ and $x^2_i$), the output of fusion gate $u_i$ is formally defined by:
\begin{equation}f_i = \sigmoid(W_1x^1_i + W_2x^2_i + b)\end{equation}
\begin{equation}u_i = f_i \odot x^1_i + (1 - f_i) \odot x^2_i\end{equation}
where $W_1$, $W_2$, and $b$ are learnable parameters. 
The fusion gate enables the model to learn how to weight the two input sequences, which solves the problem of the imbalanced importance of behavioral information and semantic information.

\paratitle{Location Profile.} 
The properties (exclude ``category'' ) of a location profile fall into two classes: \textit{word-based} and \textit{numeric-based}. 
For \textit{word-based} properties (\ie title, address, and tips), we represent each property by averaging all word embeddings inside it. 
In particular, the word embeddings of ``tips'' are weighted by TF-IDF schema. 
For \textit{numeric-based} properties (\ie tips count, visitors count and visits count), we normalize them through the whole CLP.
For \textit{category} property, we encode it as a one-hot vector. 
Finally, we concatenate representations of all properties together as the location profile representation. 

\paratitle{Manhattan Distance.} To measure the distance between user coordinates and
location coordinates in a city, we use the Manhattan distance (\ie taxicab
distance) defined by:
\begin{equation}
\label{equ:dis}
d(p, q) = \lvert p_x - q_x \rvert + \lvert p_y - q_y \rvert\end{equation}
where $p$ and $q$ are two pairs of coordinates. 
Manhattan distance is better than Euclidean distance since the road network in a city or suburb is usually orthogonal. 
Note that not every tweet has an attached user position. 
For these situations, we fix the distance value to a pre-defined constant.
Manhattan distance is also used in our post-processing, geographical pair linking, which will be
discussed in Section \ref{sec:pl}.

\paratitle{Final Prediction.} For all candidate location profiles in our mapping dictionary for a mention, we predict the matching score between the mention and each candidate. 
Formally, the final predicted location profile $p_i$ for mention $m_i$ is computed by:
\begin{equation}\label{equ:argmax}p_i = \argmax\limits_{p_j \in c(m_i)} \phi(m_i, p_j)\end{equation}
where $c(m_i)$ is the candidate set for $m_i$ and $\phi(m_i, p_j)$ is the matching score
between $m_i$ and $p_j$. 

%However, we introduce a novel strategy to the final prediction procedure to enhance
%the performance.

\section{Geographical Pair Linking}
\label{sec:pl}
In this section, we introduce Geographical Pair Linking (Geo-PL), a post-processing strategy to enhance our linking performance.

For location linking, one of the most valuable information is the geographical coherence among mentioned locations in the same tweet. First, instead of directly calling the branch name of a chain restaurant, users are more likely to mention them with another location
(usually a landmark).
Second, people may describe a route by listing the locations along the way one by one, 
which also reveals the coherence among the mentioned locations.

\begin{figure}
	\includegraphics[width=8.5cm]{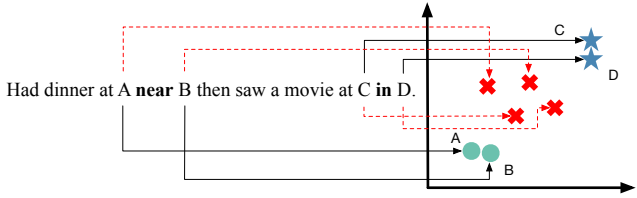}
	\caption{A wrong case of taking into account the coherence among all mentions. A, B, C and D are true POIs mentioned by a user. The incorrect predictions are marked with red dashed arrows and crosses.}
	\label{fig:pl}
\end{figure}

For geographical coherence, prior studies often measure the distances among all mentions in a tweet. 
According to our observation, geographical coherence between two locations is strong enough for location disambiguation. 
Moreover, as illustrated in Figure \ref{fig:pl}, when a user mentions four locations
with a conventional way like ``Had dinner at A near B then saw a movie at C in D,''
the geographical coherence measurement among all mentions brings obvious error. It obtains the collective minimum but disregards the strong connection between two mentions.

\begin{table*}
	\caption{Performance on Location Recognition and Location Linking.}
	\label{tab:performance}
	\begin{tabular}{l|ccc|ccc|ccc}
		\toprule
		\multirow{2}*{Methods}
		&\multicolumn{3}{c|}{Location Recognition}
		&\multicolumn{3}{c}{Location Linking}
		&\multicolumn{3}{c}{Recognition + Linking}
		\\
		&\textit{Pr} &\textit{Re} &$F_1$ &\textit{Pr} &\textit{Re} &$F_1$ &\textit{Pr} &\textit{Re} &$F_1$\\
		\midrule
		Li \textit{et al.} \cite{sigir14:li} &\textbf{0.8962} &0.7661 &0.8261 &- &- &- &- &- &-\\
		Shen \textit{et al.} \cite{kdd13:shen} &- &- &- &0.6723 &0.6349 &0.6531 &- &- &-\\
		Pipelined \cite{sigir14:li} \cite{kdd13:shen} &- &- &- &- &- &- &0.6339 &0.5635 &0.5966\\
		$\rm JoRL_L$ \cite{www16:ji} &0.8926 &0.7823 &\textbf{0.8338} &- &- &- &0.8152 &0.5952 &0.6881\\
		\ourApproach &0.8575 &\textbf{0.8091} &0.8326 &\textbf{0.8235} &\textbf{0.7778} &\textbf{0.8000} &\textbf{0.8350} &\textbf{0.6825} &\textbf{0.7511}\\
		\bottomrule
	\end{tabular}
\end{table*}

To exploit pair-wise geographical coherence, we developed \textbf{Geographical Pair Linking (Geo-PL)} algorithm based on the \textit{Pair Linking} algorithm \cite{cikm17:phan}. 
The original algorithm measures the cosine similarity between text embeddings, while we use Manhattan distance discussed in Section
\ref{subsec:linking} to measure the geographical coherence. 
Geo-PL iteratively resolves all pairs of mentions, starting from the most \textit{confident} pair. 
The confidence score of a pair 
of links $m_i \mapsto p_i$ and $m_j \mapsto p_j$ is defined by:
\begin{equation}
conf(i, j) = (1 - \beta)\frac{[\phi(m_i,p_i)+\phi(m_j,p_j)]}{2} + \beta \frac{1}{d(p_i, p_j)}
\end{equation}
where $\beta$ is a given coefficient representing the preference between the 
matching scores and the geographical coherence; $d$ is the Manhattan distance 
defined by Equation \ref{equ:dis}. In the case of $p_i = p_j$, we temporarily set 
$\beta$ to 0 when calculating. The procedure of Geo-PL (the same as Pair Linking)
is detailed in Algorithm 1. 
Note that if there is only one POI mention in a tweet, we simply predict its location profile by Equation \ref{equ:argmax}.

\begin{algorithm}
\small
	\caption{Geo-PL Algorithm}%算法名字
	\LinesNumbered %要求显示行号
	\KwIn{N mentions($m_1,...,m_N$). Mention $m_i$ has candidate profiles $\{e|e_i \in C(m_i)\}$}%输入参数
	\KwOut{$\Gamma = (p_1,...,p_N)$}%输出
	$p_i \gets null, \forall p_i \in \Gamma$\;
	\ForEach{$pair(m_i, m_j)\land m_i \not= m_j$}{
		$Q_{m_i,m_j} \gets top\_pair(m_i,C(m_i),m_j,C(m_j))$\;
		$Q.add(Q_{m_i,m_j})$\;
	}
	\While{$\exists p_i \in \Gamma, p_i = null$}{
		$(m_i, e_i, m_j, e_j)\gets most\_confident\_pair(Q)$\;
		$p_i \gets e_i$\;
		$p_j \gets e_j$\;
		\For{$k:=1 \to N \land p_k=null$}{
			$Q_{m_k,m_i} \gets top\_pair(m_k,C(m_k),m_i,\{p_i\})$\;
			$Q_{m_k,m_j} \gets top\_pair(m_k,C(m_k),m_j,\{p_j\})$\;
		}
	}
\end{algorithm}

\section{Experiment}
\subsection{Data Preparation}
We use a Singaporean national Twitter dataset released by Ji \etal \cite{www16:ji}. 
The dataset includes a CLP and a set of labeled tweets. 

326,853 Foursquare \textit{check-in tweets}, containing both formal auto-generated POI name and informal user mentions, are collected to build the CLP (including 22,414 valid location profiles) and the POI inventory (including 27,386 entries).
Location profile mapping dictionary (see Section \ref{subsec:linking}) is also constructed with check-in tweets. 
Informal mentions in check-in tweets and the indexes of  crawled profiles are used as keys and values,  respectively.
The final location profile mapping dictionary has 24,750 keys and 63,091 \textit{<key, value>}
mappings.

The dataset consists of 3,611 labeled tweets and 1,542 POI locations, 543 of which can be linked to a profile in the CLP. 
10\% of tweets have an attached user position.
These tweets are labeled by human annotators. 
All possible POI mentions are labeled with one of \textit{NPOI} (\ie not a POI mention), \textit{index} of linked location profile, or \textit{NIL} (\ie cannot be linked to a profile).

We split the 3,611 labeled tweets randomly into three subsets: 2,500 tweets for training, 211 tweets for
validation, and 900 tweets for testing.
Same POI inventory and location profile mapping dictionary used in \cite{www16:ji} are employed through all models in our experiments.

\subsection{Parameter Setting and Evaluation Metrics}
We set $\beta$, the coefficient of Geo-PL preference to 0.8 based on fine-tuning on the validation set. 
When concatenating the scalars to be the input of MLP, we duplicate them to the same number of dimensions
as the word-based properties (\ie 200 dimensions in our experiments). 
This setting prevents the scalars from being ``ignored'' during training.

We conduct experiments on location recognition subtask, location
linking subtask and the whole task, respectively. 
We adopt three metrics, Precision ($Pr$), Recall ($Re$) and $F_1$ which are widely used to evaluate NER and EL tasks. 
In particular, we use the ground-truth mentions instead of  the predicted output of the recognition module, to get the metric scores for location linking.

\subsection{Overall Comparison}
\label{subsec:overallcmp}
We compare our model with state-of-the-art solutions. 
More specifically, we compare our model with \cite{sigir14:li} and \cite{www16:ji} for recognition, and \cite{kdd13:shen} for linking, respectively. 
Note that the performance of \cite{www16:ji} on linking subtask is not provided since it is a joint model. 
For the whole task, we choose $\rm JoRL_L$ as the baseline since it is the state-of-the-art joint model in supervised learning manner. 
All models are fine-tuned on the validation set.

The result is shown in Table \ref{tab:performance}.
We observe that \ourApproach achieves the highest \textit{recall} on location recognition. 
\ourApproach beats Li \textit{et al.} \cite{sigir14:li}  and $\rm JoRL_L$ \cite{www16:ji} with relative recall improvements of $5.61\%$ and $3.43\%$, respectively. 
We contribute this to the fact that word/character embeddings enable our model to be more ``tolerant'' than prior works which
depend on hand-crafted features.
Although the precision score of \ourApproach is not good, it does not have an impact on the whole task, because the location profile mapping dictionary can filter out most false positive (FP) prediction in the linking module.

On location linking subtask, our method outperforms the work \cite{kdd13:shen} by 22.49\% on 
precision, 22.51\% on recall and 22.49\% on $F_1$.

On the whole task, our model prominently outperforms the 
state-of-the-art joint solution ($\rm JoRL_L$) on all three metrics (2.43\% on precision, 14.67\% on
recall, 9.16\% on $F_1$). Also, our model dramatically outperforms the prior state-of-the-art 
\textit{pipeline} by 31.72\% on precision, 21.12\% on recall and 25.90\% on $F_1$.

\subsection{Effect of Multi-attention for Linking}
We conduct experiments to verify the effectiveness of the multi-attention mechanism. 
Table \ref{tab:attentions} shows the linking performance with single attention from tips/title attentions and multi-attention. 
Note that the single attention approaches can only slightly improve the performance because they have ``prejudice'' which considers either behavioral information or semantic information.
By ``fusing'' this two information, \ourApproach eliminates the prejudice so it can effectively filter out noisy information.
With the multi-attention mechanism, \ourApproach improves the performance of  linking module by 4.25\% on precision, 4.26\% on recall, and 4.26\% on $F_1$, compared with the baseline.

\subsection{Effect of Geo-PL for Linking}
As discussed in Section \ref{sec:pl}, we introduce a novel post-processing strategy, named Geo-PL. 
Here, we conduct experiments to show the impact of Geo-PL for location linking.
Table \ref{tab:pl} shows the experimental results. 
Compared with no Geo-PL component, \ourApproach (with Geo-PL) significantly improves precision by 13.46\%, recall by 8.89\% and $F_1$ by 11.11\%.

\begin{table}
	\caption{Effect of Multi-attention Mechanism for Linking.}
	\label{tab:attentions}
	\begin{tabular}{l|ccc}
		\toprule
		\multirow{2}*{Attention}
		&\multicolumn{3}{c}{Location Linking}
		\\
		&\textit{Pr} &\textit{Re} &$F_1$\\
		\midrule
		Baseline (\textit{without attention}) &0.7899 &0.7460 &0.7673\\
		Single Tips Attention &0.7983 &0.7540 &0.7755\\
		Single Title Attention &0.7899 &0.7460 &0.7673\\
		\hline
		Tips + Title (\textit{multi-attention}) &0.8235 &0.7778 &0.8000\\
		\bottomrule
	\end{tabular}
\end{table}

\begin{table}
	\caption{Effect of Geo-PL for Linking.}
	\label{tab:pl}
	\begin{tabular}{l|ccc}
		\toprule
		\multirow{2}*{Strategy}
		&\multicolumn{3}{c}{Location Linking}
		\\
		&\textit{Pr} &\textit{Re} &$F_1$\\
		\midrule
		Without Geo-PL &0.7258 &0.7143 &0.7200\\
		With  Geo-PL &0.8235 &0.7778 &0.8000\\
		\bottomrule
	\end{tabular} 
\end{table}
\section{Conclusions}
In this paper, we introduce \ourApproach, the first deep neural network based pipeline to recognize 
fine-grained location mentions in tweets and link the recognized locations to location 
profiles. 
Moreover, we develop a novel post-processing strategy which can further improve location linking performance. 
Through extensive experiments, we demonstrate the effectiveness of \ourApproach against state-of-the-art solutions on a real-world Twitter dataset.
The ablation experiments show the effectiveness of multi-attention mechanism and Geo-PL strategy on location linking.

\begin{acks}
We would like to thank the anonymous reviewers for their careful reading and their many insightful comments and suggestions.
This research was supported by National Natural Science Foundation of China (No.U1636219, No.U1804263, No.61872278, No.61502344), Natural Scientific Research Program of Hubei Province (No.2017CFB\-502, No.2017CFA007), the National Key R\&D Program of China (No.2016QY01W0105, No.2016YFB0801303), and Plan for Scientific Innovation Talent of Henan Province (No.2018JR0018). Xiangyang Luo is the corresponding author.
\end{acks}
\bibliographystyle{ACM-Reference-Format}
\bibliography{sample-bibliography}


\begin{thebibliography}{36}

%%% the Web address.

\ifx \showCODEN    \undefined \def \showCODEN     #1{\unskip}     \fi
\ifx \showDOI      \undefined \def \showDOI       #1{#1}\fi
\ifx \showISBNx    \undefined \def \showISBNx     #1{\unskip}     \fi
\ifx \showISBNxiii \undefined \def \showISBNxiii  #1{\unskip}     \fi
\ifx \showISSN     \undefined \def \showISSN      #1{\unskip}     \fi
\ifx \showLCCN     \undefined \def \showLCCN      #1{\unskip}     \fi
\ifx \shownote     \undefined \def \shownote      #1{#1}          \fi
\ifx \showarticletitle \undefined \def \showarticletitle #1{#1}   \fi
\ifx \showURL      \undefined \def \showURL       {\relax}        \fi
% invisible to TeX
\providecommand\bibfield[2]{#2}
\providecommand\bibinfo[2]{#2}
\providecommand\natexlab[1]{#1}
\providecommand\showeprint[2][]{arXiv:#2}

\bibitem[\protect\citeauthoryear{Akbik, Blythe, and Vollgraf}{Akbik
  et~al\mbox{.}}{2018}]%
        {akbik2018contextual}
\bibfield{author}{\bibinfo{person}{Alan Akbik}, \bibinfo{person}{Duncan
  Blythe}, {and} \bibinfo{person}{Roland Vollgraf}.}
  \bibinfo{year}{2018}\natexlab{}.
\newblock \showarticletitle{Contextual String Embeddings for Sequence
  Labeling}. In \bibinfo{booktitle}{\emph{Proc. COLING}}.
  \bibinfo{pages}{1638--1649}.
\newblock


\bibitem[\protect\citeauthoryear{Balog}{Balog}{2018}]%
        {balog2018entity}
\bibfield{author}{\bibinfo{person}{Krisztian Balog}.}
  \bibinfo{year}{2018}\natexlab{}.
\newblock \showarticletitle{Entity-Oriented Search}.
\newblock  (\bibinfo{year}{2018}).
\newblock


\bibitem[\protect\citeauthoryear{Brown, Pietra, de~Souza, Lai, and
  Mercer}{Brown et~al\mbox{.}}{1992}]%
        {coling92:brown}
\bibfield{author}{\bibinfo{person}{Peter~F. Brown}, \bibinfo{person}{Vincent
  J.~Della Pietra}, \bibinfo{person}{Peter~V. de Souza},
  \bibinfo{person}{Jennifer~C. Lai}, {and} \bibinfo{person}{Robert~L. Mercer}.}
  \bibinfo{year}{1992}\natexlab{}.
\newblock \showarticletitle{Class-Based n-gram Models of Natural Language}.
\newblock \bibinfo{journal}{\emph{Computational Linguistics}}
  \bibinfo{volume}{18}, \bibinfo{number}{4} (\bibinfo{year}{1992}),
  \bibinfo{pages}{467--479}.
\newblock


\bibitem[\protect\citeauthoryear{Chiu and Nichols}{Chiu and Nichols}{2016}]%
        {chiu2015named}
\bibfield{author}{\bibinfo{person}{Jason~PC Chiu} {and} \bibinfo{person}{Eric
  Nichols}.} \bibinfo{year}{2016}\natexlab{}.
\newblock \showarticletitle{Named entity recognition with bidirectional
  LSTM-CNNs}. In \bibinfo{booktitle}{\emph{TACL}}, Vol.~\bibinfo{volume}{4}.
  \bibinfo{pages}{357--370}.
\newblock


\bibitem[\protect\citeauthoryear{Collobert, Weston, Bottou, Karlen,
  Kavukcuoglu, and Kuksa}{Collobert et~al\mbox{.}}{2011}]%
        {collobert2011natural}
\bibfield{author}{\bibinfo{person}{Ronan Collobert}, \bibinfo{person}{Jason
  Weston}, \bibinfo{person}{L{\'e}on Bottou}, \bibinfo{person}{Michael Karlen},
  \bibinfo{person}{Koray Kavukcuoglu}, {and} \bibinfo{person}{Pavel Kuksa}.}
  \bibinfo{year}{2011}\natexlab{}.
\newblock \showarticletitle{Natural language processing (almost) from scratch}.
\newblock \bibinfo{journal}{\emph{Journal of Machine Learning Research}}
  \bibinfo{volume}{12}, \bibinfo{number}{Aug} (\bibinfo{year}{2011}),
  \bibinfo{pages}{2493--2537}.
\newblock


\bibitem[\protect\citeauthoryear{Dyer, Ballesteros, Ling, Matthews, and
  Smith}{Dyer et~al\mbox{.}}{2015}]%
        {acl15:dyer}
\bibfield{author}{\bibinfo{person}{Chris Dyer}, \bibinfo{person}{Miguel
  Ballesteros}, \bibinfo{person}{Wang Ling}, \bibinfo{person}{Austin Matthews},
  {and} \bibinfo{person}{Noah~A. Smith}.} \bibinfo{year}{2015}\natexlab{}.
\newblock \showarticletitle{Transition-Based Dependency Parsing with Stack Long
  Short-Term Memory}. In \bibinfo{booktitle}{\emph{Proc ACL}}.
  \bibinfo{pages}{334--343}.
\newblock


\bibitem[\protect\citeauthoryear{Espinoza, Persson, Sandin, Nystr{\"o}m,
  Cacciatore, and Bylund}{Espinoza et~al\mbox{.}}{2001}]%
        {espinoza2001geonotes}
\bibfield{author}{\bibinfo{person}{Fredrik Espinoza}, \bibinfo{person}{Per
  Persson}, \bibinfo{person}{Anna Sandin}, \bibinfo{person}{Hanna Nystr{\"o}m},
  \bibinfo{person}{Elenor Cacciatore}, {and} \bibinfo{person}{Markus Bylund}.}
  \bibinfo{year}{2001}\natexlab{}.
\newblock \showarticletitle{Geonotes: Social and navigational aspects of
  location-based information systems}. In \bibinfo{booktitle}{\emph{Proc.
  UBICOMP}}. \bibinfo{pages}{2--17}.
\newblock


\bibitem[\protect\citeauthoryear{Goldberg}{Goldberg}{2017}]%
        {goldberg2017neural}
\bibfield{author}{\bibinfo{person}{Yoav Goldberg}.}
  \bibinfo{year}{2017}\natexlab{}.
\newblock \showarticletitle{Neural network methods for natural language
  processing}.
\newblock \bibinfo{journal}{\emph{Synthesis Lectures on Human Language
  Technologies}} \bibinfo{volume}{10}, \bibinfo{number}{1}
  (\bibinfo{year}{2017}), \bibinfo{pages}{1--309}.
\newblock


\bibitem[\protect\citeauthoryear{Guo, Chang, and Kiciman}{Guo
  et~al\mbox{.}}{2013}]%
        {naacl13:guo}
\bibfield{author}{\bibinfo{person}{Stephen Guo}, \bibinfo{person}{Ming{-}Wei
  Chang}, {and} \bibinfo{person}{Emre Kiciman}.}
  \bibinfo{year}{2013}\natexlab{}.
\newblock \showarticletitle{To Link or Not to Link? {A} Study on End-to-End
  Tweet Entity Linking}. In \bibinfo{booktitle}{\emph{Proc. NAACL-HLT}}.
  \bibinfo{pages}{1020--1030}.
\newblock


\bibitem[\protect\citeauthoryear{Han, Sun, Cong, Zhao, Ji, and Phan}{Han
  et~al\mbox{.}}{2018}]%
        {han2018linking}
\bibfield{author}{\bibinfo{person}{Jialong Han}, \bibinfo{person}{Aixin Sun},
  \bibinfo{person}{Gao Cong}, \bibinfo{person}{Wayne~Xin Zhao},
  \bibinfo{person}{Zongcheng Ji}, {and} \bibinfo{person}{Minh~C Phan}.}
  \bibinfo{year}{2018}\natexlab{}.
\newblock \showarticletitle{Linking Fine-Grained Locations in User Comments}.
\newblock \bibinfo{journal}{\emph{IEEE Transactions on Knowledge and Data
  Engineering}} \bibinfo{volume}{30}, \bibinfo{number}{1}
  (\bibinfo{year}{2018}), \bibinfo{pages}{59--72}.
\newblock


\bibitem[\protect\citeauthoryear{Han, Sun, and Zhao}{Han et~al\mbox{.}}{2011}]%
        {sigir11:han}
\bibfield{author}{\bibinfo{person}{Xianpei Han}, \bibinfo{person}{Le Sun},
  {and} \bibinfo{person}{Jun Zhao}.} \bibinfo{year}{2011}\natexlab{}.
\newblock \showarticletitle{Collective entity linking in web text: a
  graph-based method}. In \bibinfo{booktitle}{\emph{Proc. SIGIR}}.
  \bibinfo{pages}{765--774}.
\newblock


\bibitem[\protect\citeauthoryear{Hoffart, Yosef, Bordino, F{\"{u}}rstenau,
  Pinkal, Spaniol, Taneva, Thater, and Weikum}{Hoffart et~al\mbox{.}}{2011}]%
        {emnlp11:hoffart}
\bibfield{author}{\bibinfo{person}{Johannes Hoffart},
  \bibinfo{person}{Mohamed~Amir Yosef}, \bibinfo{person}{Ilaria Bordino},
  \bibinfo{person}{Hagen F{\"{u}}rstenau}, \bibinfo{person}{Manfred Pinkal},
  \bibinfo{person}{Marc Spaniol}, \bibinfo{person}{Bilyana Taneva},
  \bibinfo{person}{Stefan Thater}, {and} \bibinfo{person}{Gerhard Weikum}.}
  \bibinfo{year}{2011}\natexlab{}.
\newblock \showarticletitle{Robust Disambiguation of Named Entities in Text}.
  In \bibinfo{booktitle}{\emph{Proc. EMNLP}}. \bibinfo{pages}{782--792}.
\newblock


\bibitem[\protect\citeauthoryear{Huang, Xu, and Yu}{Huang
  et~al\mbox{.}}{2015}]%
        {huang2015bidirectional}
\bibfield{author}{\bibinfo{person}{Zhiheng Huang}, \bibinfo{person}{Wei Xu},
  {and} \bibinfo{person}{Kai Yu}.} \bibinfo{year}{2015}\natexlab{}.
\newblock \showarticletitle{Bidirectional LSTM-CRF models for sequence
  tagging}.
\newblock \bibinfo{journal}{\emph{arXiv preprint arXiv:1508.01991}}
  (\bibinfo{year}{2015}).
\newblock


\bibitem[\protect\citeauthoryear{Ji, Sun, Cong, and Han}{Ji
  et~al\mbox{.}}{2016}]%
        {www16:ji}
\bibfield{author}{\bibinfo{person}{Zongcheng Ji}, \bibinfo{person}{Aixin Sun},
  \bibinfo{person}{Gao Cong}, {and} \bibinfo{person}{Jialong Han}.}
  \bibinfo{year}{2016}\natexlab{}.
\newblock \showarticletitle{Joint Recognition and Linking of Fine-Grained
  Locations from Tweets}. In \bibinfo{booktitle}{\emph{Proc. WWW}}.
  \bibinfo{pages}{1271--1281}.
\newblock


\bibitem[\protect\citeauthoryear{Kulkarni, Singh, Ramakrishnan, and
  Chakrabarti}{Kulkarni et~al\mbox{.}}{2009}]%
        {kdd09:kulkarni}
\bibfield{author}{\bibinfo{person}{Sayali Kulkarni}, \bibinfo{person}{Amit
  Singh}, \bibinfo{person}{Ganesh Ramakrishnan}, {and} \bibinfo{person}{Soumen
  Chakrabarti}.} \bibinfo{year}{2009}\natexlab{}.
\newblock \showarticletitle{Collective annotation of Wikipedia entities in web
  text}. In \bibinfo{booktitle}{\emph{Proc. SIGKDD}}.
  \bibinfo{pages}{457--466}.
\newblock


\bibitem[\protect\citeauthoryear{Li and Sun}{Li and Sun}{2014}]%
        {sigir14:li}
\bibfield{author}{\bibinfo{person}{Chenliang Li} {and} \bibinfo{person}{Aixin
  Sun}.} \bibinfo{year}{2014}\natexlab{}.
\newblock \showarticletitle{Fine-grained location extraction from tweets with
  temporal awareness}. In \bibinfo{booktitle}{\emph{Proc. SIGIR}}.
  \bibinfo{pages}{43--52}.
\newblock


\bibitem[\protect\citeauthoryear{Li, Luo, Ding, Tang, Sun, Dai, Du, Zhang, and
  Kong}{Li et~al\mbox{.}}{2017}]%
        {jasist17:li}
\bibfield{author}{\bibinfo{person}{Daifeng Li}, \bibinfo{person}{Zhipeng Luo},
  \bibinfo{person}{Ying Ding}, \bibinfo{person}{Jie Tang},
  \bibinfo{person}{Gordon~Guo{-}Zheng Sun}, \bibinfo{person}{Xiaowen Dai},
  \bibinfo{person}{John Du}, \bibinfo{person}{Jingwei Zhang}, {and}
  \bibinfo{person}{Shoubin Kong}.} \bibinfo{year}{2017}\natexlab{}.
\newblock \showarticletitle{User-level microblogging recommendation
  incorporating social influence}.
\newblock \bibinfo{journal}{\emph{Journal of the Association for Information
  Science and Technology}} \bibinfo{volume}{68}, \bibinfo{number}{3}
  (\bibinfo{year}{2017}), \bibinfo{pages}{553--568}.
\newblock


\bibitem[\protect\citeauthoryear{Li, Hu, Feng, and Tan}{Li
  et~al\mbox{.}}{2014}]%
        {icde14:li}
\bibfield{author}{\bibinfo{person}{Guoliang Li}, \bibinfo{person}{Jun Hu},
  \bibinfo{person}{Jianhua Feng}, {and} \bibinfo{person}{Kian{-}Lee Tan}.}
  \bibinfo{year}{2014}\natexlab{}.
\newblock \showarticletitle{Effective location identification from microblogs}.
  In \bibinfo{booktitle}{\emph{Proc. ICDE}}. \bibinfo{pages}{880--891}.
\newblock


\bibitem[\protect\citeauthoryear{Li, Sun, and Joty}{Li et~al\mbox{.}}{2018}]%
        {ijcai2018-579}
\bibfield{author}{\bibinfo{person}{Jing Li}, \bibinfo{person}{Aixin Sun}, {and}
  \bibinfo{person}{Shafiq Joty}.} \bibinfo{year}{2018}\natexlab{}.
\newblock \showarticletitle{SegBot: A Generic Neural Text Segmentation Model
  with Pointer Network}. In \bibinfo{booktitle}{\emph{Proc. IJCAI}}.
  \bibinfo{pages}{4166--4172}.
\newblock


\bibitem[\protect\citeauthoryear{Lingad, Karimi, and Yin}{Lingad
  et~al\mbox{.}}{2013}]%
        {www13:lingad}
\bibfield{author}{\bibinfo{person}{John Lingad}, \bibinfo{person}{Sarvnaz
  Karimi}, {and} \bibinfo{person}{Jie Yin}.} \bibinfo{year}{2013}\natexlab{}.
\newblock \showarticletitle{Location extraction from disaster-related
  microblogs}. In \bibinfo{booktitle}{\emph{Proc. WWW}}.
  \bibinfo{pages}{1017--1020}.
\newblock


\bibitem[\protect\citeauthoryear{Malmasi and Dras}{Malmasi and Dras}{2015}]%
        {pacling15:malmasi}
\bibfield{author}{\bibinfo{person}{Shervin Malmasi} {and} \bibinfo{person}{Mark
  Dras}.} \bibinfo{year}{2015}\natexlab{}.
\newblock \showarticletitle{Location Mention Detection in Tweets and
  Microblogs}. In \bibinfo{booktitle}{\emph{Proc. PACLING}}.
  \bibinfo{pages}{123--134}.
\newblock


\bibitem[\protect\citeauthoryear{Oh and Xu}{Oh and Xu}{2003}]%
        {oh2003effects}
\bibfield{author}{\bibinfo{person}{Lih-Bin Oh} {and} \bibinfo{person}{Heng
  Xu}.} \bibinfo{year}{2003}\natexlab{}.
\newblock \showarticletitle{Effects of multimedia on mobile consumer behavior:
  An empirical study of location-aware advertising}.
\newblock \bibinfo{journal}{\emph{Proc. ICiS}} (\bibinfo{year}{2003}),
  \bibinfo{pages}{56}.
\newblock


\bibitem[\protect\citeauthoryear{Pennington, Socher, and Manning}{Pennington
  et~al\mbox{.}}{2014}]%
        {emnlp14:pennington}
\bibfield{author}{\bibinfo{person}{Jeffrey Pennington},
  \bibinfo{person}{Richard Socher}, {and} \bibinfo{person}{Christopher~D.
  Manning}.} \bibinfo{year}{2014}\natexlab{}.
\newblock \showarticletitle{Glove: Global Vectors for Word Representation}. In
  \bibinfo{booktitle}{\emph{Proc. EMNLP}}. \bibinfo{pages}{1532--1543}.
\newblock


\bibitem[\protect\citeauthoryear{Peters, Neumann, Iyyer, Gardner, Clark, Lee,
  and Zettlemoyer}{Peters et~al\mbox{.}}{2018}]%
        {peters2018deep}
\bibfield{author}{\bibinfo{person}{Matthew~E Peters}, \bibinfo{person}{Mark
  Neumann}, \bibinfo{person}{Mohit Iyyer}, \bibinfo{person}{Matt Gardner},
  \bibinfo{person}{Christopher Clark}, \bibinfo{person}{Kenton Lee}, {and}
  \bibinfo{person}{Luke Zettlemoyer}.} \bibinfo{year}{2018}\natexlab{}.
\newblock \showarticletitle{Deep contextualized word representations}. In
  \bibinfo{booktitle}{\emph{Proc. NAACL-HLT}}. \bibinfo{pages}{2227--2237}.
\newblock


\bibitem[\protect\citeauthoryear{Phan, Sun, Tay, Han, and Li}{Phan
  et~al\mbox{.}}{2017}]%
        {cikm17:phan}
\bibfield{author}{\bibinfo{person}{Minh~C. Phan}, \bibinfo{person}{Aixin Sun},
  \bibinfo{person}{Yi Tay}, \bibinfo{person}{Jialong Han}, {and}
  \bibinfo{person}{Chenliang Li}.} \bibinfo{year}{2017}\natexlab{}.
\newblock \showarticletitle{NeuPL: Attention-based Semantic Matching and
  Pair-Linking for Entity Disambiguation}. In \bibinfo{booktitle}{\emph{Proc.
  CIKM}}. \bibinfo{pages}{1667--1676}.
\newblock


\bibitem[\protect\citeauthoryear{Rae, Murdock, Popescu, and Bouchard}{Rae
  et~al\mbox{.}}{2012}]%
        {rae2012mining}
\bibfield{author}{\bibinfo{person}{Adam Rae}, \bibinfo{person}{Vanessa
  Murdock}, \bibinfo{person}{Adrian Popescu}, {and} \bibinfo{person}{Hugues
  Bouchard}.} \bibinfo{year}{2012}\natexlab{}.
\newblock \showarticletitle{Mining the web for points of interest}. In
  \bibinfo{booktitle}{\emph{Proc. SIGIR}}. \bibinfo{pages}{711--720}.
\newblock


\bibitem[\protect\citeauthoryear{Ratinov and Roth}{Ratinov and Roth}{2009}]%
        {conll09:ratinov}
\bibfield{author}{\bibinfo{person}{Lev{-}Arie Ratinov} {and}
  \bibinfo{person}{Dan Roth}.} \bibinfo{year}{2009}\natexlab{}.
\newblock \showarticletitle{Design Challenges and Misconceptions in Named
  Entity Recognition}. In \bibinfo{booktitle}{\emph{Proc. CoNLL}}.
  \bibinfo{pages}{147--155}.
\newblock


\bibitem[\protect\citeauthoryear{Schiller and Voisard}{Schiller and
  Voisard}{2004}]%
        {schiller2004location}
\bibfield{author}{\bibinfo{person}{Jochen Schiller} {and}
  \bibinfo{person}{Agn{\`e}s Voisard}.} \bibinfo{year}{2004}\natexlab{}.
\newblock \bibinfo{booktitle}{\emph{Location-based services}}.
\newblock \bibinfo{publisher}{Elsevier}.
\newblock


\bibitem[\protect\citeauthoryear{Shen, Zhou, Long, Jiang, Pan, and Zhang}{Shen
  et~al\mbox{.}}{2018}]%
        {aaai18:shen}
\bibfield{author}{\bibinfo{person}{Tao Shen}, \bibinfo{person}{Tianyi Zhou},
  \bibinfo{person}{Guodong Long}, \bibinfo{person}{Jing Jiang},
  \bibinfo{person}{Shirui Pan}, {and} \bibinfo{person}{Chengqi Zhang}.}
  \bibinfo{year}{2018}\natexlab{}.
\newblock \showarticletitle{DiSAN: Directional Self-Attention Network for
  RNN/CNN-Free Language Understanding}. In \bibinfo{booktitle}{\emph{Proc.
  AAAI}}.
\newblock


\bibitem[\protect\citeauthoryear{Shen, Wang, Luo, and Wang}{Shen
  et~al\mbox{.}}{2013}]%
        {kdd13:shen}
\bibfield{author}{\bibinfo{person}{Wei Shen}, \bibinfo{person}{Jianyong Wang},
  \bibinfo{person}{Ping Luo}, {and} \bibinfo{person}{Min Wang}.}
  \bibinfo{year}{2013}\natexlab{}.
\newblock \showarticletitle{Linking named entities in Tweets with knowledge
  base via user interest modeling}. In \bibinfo{booktitle}{\emph{Proc.
  SIGKDD}}. \bibinfo{pages}{68--76}.
\newblock


\bibitem[\protect\citeauthoryear{Sun, Lin, Tang, Yang, Ji, and Wang}{Sun
  et~al\mbox{.}}{2015}]%
        {ijcai15:sun}
\bibfield{author}{\bibinfo{person}{Yaming Sun}, \bibinfo{person}{Lei Lin},
  \bibinfo{person}{Duyu Tang}, \bibinfo{person}{Nan Yang},
  \bibinfo{person}{Zhenzhou Ji}, {and} \bibinfo{person}{Xiaolong Wang}.}
  \bibinfo{year}{2015}\natexlab{}.
\newblock \showarticletitle{Modeling Mention, Context and Entity with Neural
  Networks for Entity Disambiguation}. In \bibinfo{booktitle}{\emph{Proc.
  IJCAI}}. \bibinfo{pages}{1333--1339}.
\newblock


\bibitem[\protect\citeauthoryear{Tsochantaridis, Joachims, Hofmann, and
  Altun}{Tsochantaridis et~al\mbox{.}}{2005}]%
        {jmlr05:tsochantaridis}
\bibfield{author}{\bibinfo{person}{Ioannis Tsochantaridis},
  \bibinfo{person}{Thorsten Joachims}, \bibinfo{person}{Thomas Hofmann}, {and}
  \bibinfo{person}{Yasemin Altun}.} \bibinfo{year}{2005}\natexlab{}.
\newblock \showarticletitle{Large Margin Methods for Structured and
  Interdependent Output Variables}.
\newblock \bibinfo{journal}{\emph{Journal of Machine Learning Research}}
  \bibinfo{volume}{6} (\bibinfo{year}{2005}), \bibinfo{pages}{1453--1484}.
\newblock


\bibitem[\protect\citeauthoryear{Yang, Salakhutdinov, and Cohen}{Yang
  et~al\mbox{.}}{2016}]%
        {yang2016multi}
\bibfield{author}{\bibinfo{person}{Zhilin Yang}, \bibinfo{person}{Ruslan
  Salakhutdinov}, {and} \bibinfo{person}{William Cohen}.}
  \bibinfo{year}{2016}\natexlab{}.
\newblock \showarticletitle{Multi-task cross-lingual sequence tagging from
  scratch}.
\newblock \bibinfo{journal}{\emph{arXiv preprint arXiv:1603.06270}}
  (\bibinfo{year}{2016}).
\newblock


\bibitem[\protect\citeauthoryear{Zhang and Gelernter}{Zhang and
  Gelernter}{2014}]%
        {josis14:zhang}
\bibfield{author}{\bibinfo{person}{Wei Zhang} {and} \bibinfo{person}{Judith
  Gelernter}.} \bibinfo{year}{2014}\natexlab{}.
\newblock \showarticletitle{Geocoding location expressions in Twitter messages:
  {A} preference learning method}.
\newblock \bibinfo{journal}{\emph{J. Spatial Information Science}}
  \bibinfo{volume}{9}, \bibinfo{number}{1} (\bibinfo{year}{2014}),
  \bibinfo{pages}{37--70}.
\newblock


\bibitem[\protect\citeauthoryear{Zhang and Clark}{Zhang and Clark}{2008}]%
        {acl08:zhang}
\bibfield{author}{\bibinfo{person}{Yue Zhang} {and} \bibinfo{person}{Stephen
  Clark}.} \bibinfo{year}{2008}\natexlab{}.
\newblock \showarticletitle{Joint Word Segmentation and {POS} Tagging Using a
  Single Perceptron}. In \bibinfo{booktitle}{\emph{Proc. ACL}}.
  \bibinfo{pages}{888--896}.
\newblock


\bibitem[\protect\citeauthoryear{Zheng, Han, and Sun}{Zheng
  et~al\mbox{.}}{2018}]%
        {tkde18:zheng}
\bibfield{author}{\bibinfo{person}{Xin Zheng}, \bibinfo{person}{Jialong Han},
  {and} \bibinfo{person}{Aixin Sun}.} \bibinfo{year}{2018}\natexlab{}.
\newblock \showarticletitle{A Survey of Location Prediction on Twitter}.
\newblock \bibinfo{journal}{\emph{{IEEE} Trans. Knowl. Data Eng.}}
  \bibinfo{volume}{30}, \bibinfo{number}{9} (\bibinfo{year}{2018}),
  \bibinfo{pages}{1652--1671}.
\newblock


\end{thebibliography}

\end{document}